\documentclass[10pt,twocolumn,letterpaper]{article}

\usepackage{wacv}      

\usepackage{graphicx}
\usepackage{amsmath}
\usepackage{amssymb}
\usepackage{booktabs}
\usepackage{colortbl}
\usepackage{makecell}
\usepackage[table]{xcolor}

\def\hmath$#1${\texorpdfstring{{\rmfamily\textit{#1}}}{#1}}
\usepackage{multirow}
\usepackage[pagebackref,breaklinks,colorlinks]{hyperref}

\usepackage{etoolbox,lipsum}
\usepackage[capitalize]{cleveref}
\crefname{section}{Sec.}{Secs.}
\Crefname{section}{Section}{Sections}
\Crefname{table}{Table}{Tables}
\crefname{table}{Tab.}{Tabs.}

\def\wacvPaperID{1767} 
\def\confName{WACV}
\def\confYear{2025}

\begin{document}

\title{Foundation Models and Adaptive Feature Selection: A Synergistic Approach to Video Question Answering}

\author{Sai Bhargav Rongali\\
Indian Institute of Technology Bombay, India\\
{\tt\small rongalisaibhargav002@gmail.com}
\and
Mohamad Hassan N C\\
Indian Institute of Technology Bombay, India\\
{\tt\small  mohdhassannc@gmail.com}
\and 
Ankit Jha\\
LNMIIT, Jaipur, India\\
{\tt\small ankitjha16@gmail.com}
\and 
Neha Bhargava\\
Fractal AI Research, India\\
{\tt\small neha.bhargava@fractal.ai}
\and 
Saurabh Prasad\\
Univesity of Houston\\
{\tt\small saurabh.prasad@ieee.org}
\and
Biplab Banerjee\\
Indian Institute of Technology Bombay, India\\
{\tt\small  getbiplab@gmail.com}
}
\twocolumn[{%
\renewcommand\twocolumn[1][]{#1}%
\maketitle
\begin{center}
\vspace{-0.8cm}
    \centering
    \captionsetup{type=figure}
    \includegraphics[width=\textwidth]{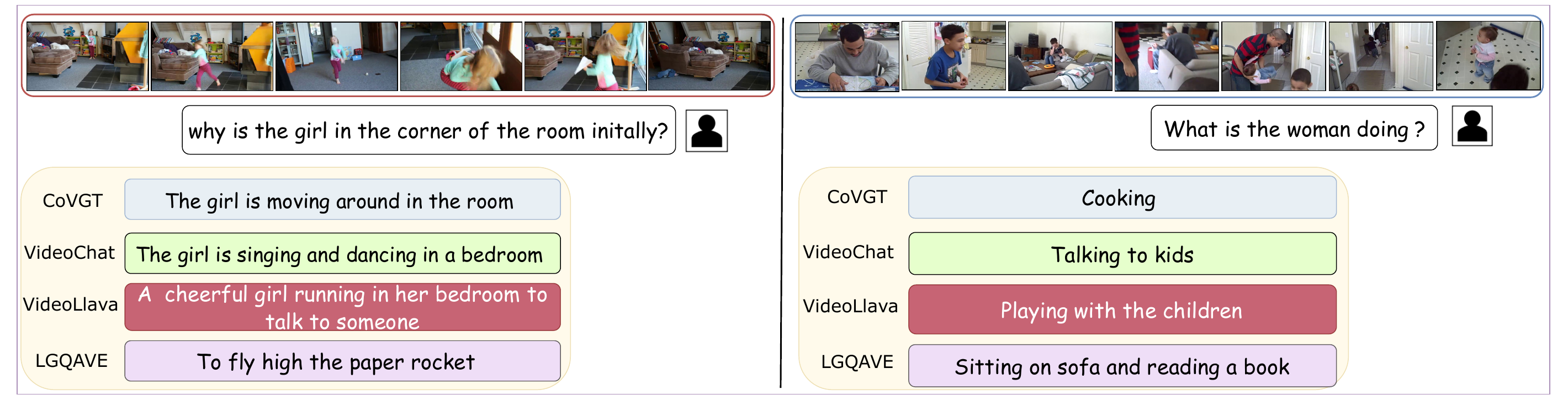}
    \vspace{-0.8cm}
    \captionof{figure}{\textbf{Qualitative analysis of LGQAVE}. We present the answers produced by various state-of-the-art VideoQA models in response to a specific question paired with a sequence of frames from a given video in the NextQA \cite{nextqa} dataset. Our findings indicate that the answers generated by our LGQAVE model are notably more direct and precise in their semantic content. \label{fig:teaser}}
\end{center}%
}]

\begin{abstract}
\vspace{-5mm}
This paper tackles the intricate challenge of video question-answering (VideoQA). Despite notable progress, current methods fall short of effectively integrating questions with video frames and semantic object-level abstractions to create question-aware video representations. We introduce \textbf{L}ocal - \textbf{G}lobal \textbf{Q}uestion \textbf{A}ware \textbf{V}ideo \textbf{E}mbedding (LGQAVE), which incorporates three major innovations to integrate multi-modal knowledge better and emphasize semantic visual concepts relevant to specific questions.
LGQAVE moves beyond traditional ad-hoc frame sampling by utilizing a cross-attention mechanism that precisely identifies the most relevant frames concerning the questions. It captures the dynamics of objects within these frames using distinct graphs, grounding them in question semantics with the miniGPT model. These graphs are processed by a question-aware dynamic graph transformer (Q-DGT), which refines the outputs to develop nuanced global and local video representations. An additional cross-attention module integrates these local and global embeddings to generate the final video embeddings, which a language model uses to generate answers.
 Extensive evaluations across multiple benchmarks demonstrate that LGQAVE significantly outperforms existing models in delivering accurate multi-choice and open-ended answers.
\end{abstract}

\section{Introduction}
Over the last decade, Video Question Answering (VideoQA) has evolved into a vital multidisciplinary field combining computer vision and natural language processing \cite{v1, v2}. Despite advances, accurately interpreting video semantics relative to queries remains challenging, primarily due to the complex interplay between video content and questions, keeping VideoQA at the forefront of research demands.
Current models focus on capturing spatiotemporal dynamics and aligning them with questions to derive answers \cite{v3, v4}, yet often require extensive dataset training and are prone to dataset biases, as frame selection is not guided by language. Recent advancements have moved beyond simple feature summarization, constructing scene and temporal graphs to depict object-level interactions \cite{CoVGT}. However, while adept at handling broad context queries, these approaches often miss the finer details necessary to analyze specific object interactions at the frame level.

Recently, the adoption of foundation models \cite{madan2024foundation} (\eg LlamaVid \cite{v5}) has significantly improved performance in several video comprehension tasks. While these models demonstrate superior performance, they face a specific challenge: they analyze all video frames indiscriminately, regardless of their relevance to the posed questions. Some studies have explored the paradigm of language-driven frame selection in contexts other than VideoQA \cite{v7, v9}. However, these approaches typically involve a complicated multi-stage pipeline, rely on secondary information sources such as image-based foundation models, or pose a multi-objective optimization framework, thus overburdening the entire process. Despite advancements in the frame selection stage, recent findings \cite{v10, CoVGT} indicate that the outcomes of modern multi-modal foundation models for VideoQA are heavily biased towards the language cues, emphasizing the importance of the fundamental question: \textit{To what extent are VideoQA outcomes relevant to the video contents?} Our research aims to precisely identify relevant video contents, both at the coarse and fine scales, guided by the given question semantics, for VideoQA.

\noindent \textbf{Our solution:} We introduce a unified solution, LGQAVE, for VideoQA with three principal novelties. 

Our approach incorporates a learnable cross-attention module for \textbf{question-aware video frame selection}, which dynamically associates the question prompt with frame-level visual embeddings. This is achieved by applying a threshold to the cross-attention scores, enabling the precise isolation of video frames that are semantically aligned with the question. This method circumvents the complexities inherent in multi-stage frame selection pipelines and can be effortlessly integrated into any VideoQA system.

Looking forward, we propose the construction of spatial graphs for frames identified as most pertinent to the questions, termed as \textbf{question-aware local object selection and their interaction modeling}. This task is approached as a question-guided visual grounding, avoiding traditional object detection frameworks. To this end, we employ the miniGPT4 model \cite{minigpt4} to process the questions alongside the selected frames, generating bounding box coordinates for the relevant objects.

Subsequently, frame-specific spatial graphs are constructed by considering the detected bounding boxes as the nodes and defining pairwise connections. When integrated with masked question embeddings, these graphs are fed into the Dynamic Graph Transformer (Q-DGT) model \cite{CoVGT}, which further refines the embeddings spatially and temporally, enhancing the semantic coherence between the visual content and the question context.

In our final step, we aim to derive both \textbf{local and global video representations} from the outputs of Q-DGT to effectively address both long-video level and fine-grained frame-level questions. This is achieved by refining the global video representation through a query-key-value-based cross-attention mechanism, utilizing localized frame-level graph embeddings for answer generation (Fig. \ref{fig:teaser}).
 Our significant contributions are summarized as,

\noindent \textbf{[-]} We introduce LGQAVE, an innovative model for Video QA that enhances the extraction of local and global video features, thoroughly guided by the question semantics. 

\noindent \textbf{[-]} Our approach begins with cross-attention for question-aware frame selection, followed by using miniGPT4 for visual grounding to establish object interaction graphs based on the posed question. We then intuitively obtain the video representations through the Q-DGT module.

\noindent \textbf{[-]} We showcase the performance of LGQAVE on distinct VideoQA tasks and ablate the model rigorously. We observe steady improvements of 2-6\% on average.

\section{Related Works}
\noindent\textbf{Video question answering (VideoQA):}
Traditional VideoQA methods have primarily used video encoders on sparse frames \cite{jin2024,7485869} or short segments \cite{10.1145/3323873.332505}, which struggle with spatiotemporal interactions and object compositionality \cite{7780459}, leading to suboptimal performance in reasoning tasks. Although cross-modal matching \cite{buch2022,chu2023,chu2024} and memory-based approaches \cite{clip4clipempiricalstudyclip,dosovitskiy2021} have improved video content extraction, they rely heavily on frame-level or clip-level representations, which are often inadequate for detailed object relation reasoning. Advances in graph-based methods have facilitated object-level rationale; however, these methods tend to use either unified graphs that do not effectively differentiate spatial from temporal relations or static graphs that ignore temporal dynamics \cite{CoVGT}.

Transformers have significantly advanced the field of VideoQA. Models developed from datasets such as HowTo100M \cite{frozenclipmodelsefficient} employ proxy tasks like masked language modeling \cite{huang2024} and specific supervisions, such as future utterance prediction \cite{xu2021}, to enhance performance. Despite outperforming traditional models \cite{yang2022learninganswer,zhu2020actbert}, transformer-based systems often focus on recognition or provide only shallow descriptions, struggling with visual relation reasoning due to noisy data and the limited scope of instructional videos \cite{chu2023}. Recent methods leveraging open-domain vision-text data face challenges with temporal relations and high operational costs. Despite their scalability, user-generated data can lead to overfitting. Large language models like BLIP-2 and MiniGPT-4 extend to video but encounter efficiency issues. Innovations such as MobileVLM and LLaMA-Vid have improved feature representation, and graphical models now effectively integrate both global and local features for enhanced dynamic reasoning \cite{blip2,minigpt4,liu2024,maaz2024,videollama}. \textit{Our LGQAVE model advances beyond existing approaches by integrating visual and linguistic synergies at multiple scales. This integration enables the extraction of both global and local perceptions of video content, effectively addressing various queries.}

\noindent\textbf{Graphs in VideoQA:} Early VideoQA models such as TGIF-QA \cite{tgifqa} and MSVD-QA \cite{8451103} targeted specific actions and objects in video clips, leveraging spatio-temporal features to generate responses. Subsequent advancements led to more sophisticated models like HME-VideoQA \cite{v3} and Co-Mem \cite{gao2018}, which utilize hierarchical memory networks and co-attentional frameworks to capture dynamic interactions within videos more effectively. HME-VideoQA builds hierarchical graphs by representing different levels of video granularity to capture temporal relationships. CoMem creates graphs through collaborative memory, linking video frames and question embeddings. Additionally, graph-based methods have proven effective for detailed visual understanding by representing video objects as graph structures. For example, LLaVA \cite{liu2024} enhances VLM performance by identifying objects pertinent to specific questions and constructing corresponding graphs. Conversely, the Contrastive Video Graph Transformer (CoVGT) \cite{CoVGT} excels in providing global representations by focusing on the overall video content. They take all the objects that are present in a frame and form a graph, yet it falls short in local representations and lacks question-specific conditioning.

Despite these advancements, processing all video frames remains computationally expensive. Current methods focus on spatio-temporal dynamics and semantic alignment, yet they often manage vast amounts of data, leading to overlooking redundant content. Essential visual cues may be neglected, diminishing the accuracy of video interpretation.



\noindent\textbf{Vision-language models (VLMs):} Multimodal learning outperforms unimodal methods in tasks that require visual-semantic integration, such as image and text bridging. Recent developments have introduced foundation models like CLIP \cite{clip}, FLORENCE \cite{florence}, and ALIGN \cite{jia2021scaling}, which are particularly effective in these multimodal contexts. These models harness large-scale image-text pairs to tackle a variety of tasks in the CV/NLP domains, including zero-shot classification, object detection, image captioning, and VQA, to name a few. Despite their efficacy with still images, these models face challenges with long video sequences, primarily due to the extensive number of tokens required to represent each frame.

Models such as CLIP and ALIGN have proven effective in video recognition \cite{promptvis,frozenclipmodelsefficient,expanding,pan2022} and video-text retrieval \cite{uatvr,clip4clipempiricalstudyclip}. However, they encounter difficulties in accurately capturing interactions between video content and labels. Innovative models like Flamingo \cite{flamingo} and BLIP-2 \cite{blip2} utilize web-scale image-text pairs, while InstructBLIP \cite{instructblip} and MiniGPT-4 \cite{minigpt4} leverage high-quality instructional data sources. Methods such as Video-LLaMA \cite{videollama} and VideoGPT \cite{videogpt} incorporate spatial and temporal pooling to overcome computational hurdles associated with long videos. LLaMA-VID \cite{v5} adopts a dual-token strategy to enhance the processing of long sequences. \textit{In contrast, \textit{LGQAVE} is designed to systematically utilize question guidance for frame selection and the modeling of relevant objects within and across frames. This approach aims to minimize redundancy and irrelevance in video features, thereby enhancing the efficiency and accuracy of VideoQA.}

\begin{figure*}[hbtp]
    \centering
    
        \includegraphics[width=\textwidth]{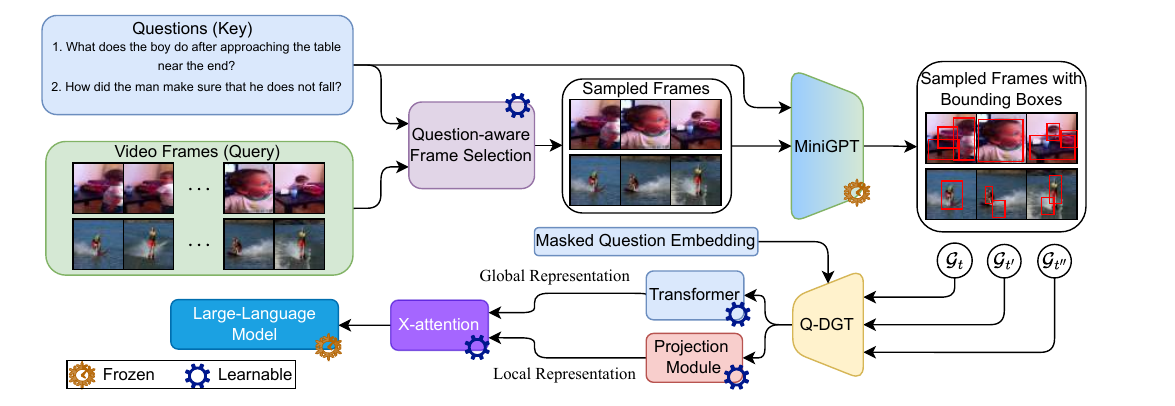}
    \caption{\textbf{Schematic of the model diagram for LGQAVE}. Given a question and its corresponding video, our process begins with a question-aware frame sampling module that identifies the pertinent frames from the video. Subsequently, a miniGPT4-based visual grounding module constructs object relation graphs from these selected frames. The Q-DGT module then processes these graphs along with masked question embeddings to produce local and global video representations. A cross-attention module further refines the global features by incorporating contextual knowledge from the local features. Finally, a language model-based answer generator utilizes these refined features to predict the answers.}
    \label{fig:model}
\vspace{-0.5cm}
\end{figure*}

\section{Proposed Methodology} 
In this section, we define the problem and outline the objectives for the LGQAVE framework. We consider a dataset \( \mathcal{D} \) that consists of video sequences \( V \), questions \( Q \), and corresponding labeled answers \( A \). The primary objective is to learn a mapping function \( \phi: (V, Q) \rightarrow A \) that accurately predicts the correct answer \( A_i \) for each given question \( Q_i \) associated with a video \( V_i \).

To accomplish this, LGQAVE is structured into four key components (Fig. \ref{fig:model}):
 \textbf{a. Question-driven frame selection module}---identifies the most relevant video frames, thereby reducing redundancy at the frame level.
\textbf{b. Frame-centric object graph construction}---emphasizes the important objects and their interactions within the selected frames through visual grounding, minimizing redundancy at a finer level.
\textbf{c. Question-aware dynamic graph transformer (Q-DGT)}---facilitates effective selection and fusion of local and global video features.
\textbf{d. Answer prediction module}---generates accurate answers based on the enriched video and question representations.
 These components, detailed below, collaboratively leverage the semantics of the questions to ensure a discriminative and contextually rich embedding space. The variables are summarized in the \textcolor{blue}{supplementary materials}.

 \subsection{Question-aware video frame selection} 

In VideoQA, processing every video frame in a sequence is both computationally intensive and time-consuming, often leading to redundancy when dealing with frame splits. To tackle these issues, our LGQAVE framework incorporates a novel frame selection module designed to sample question-aware, key video frames from video-question pairs. This is achieved by utilizing a cross-attention mechanism to calculate relevance scores between the question tokens and video frames, ensuring that only the most pertinent frames are selected.

Mathematically, we denote the video frame at the \( t^{\text{th}} \) time step for the \( i^{\text{th}} \) instance as \( V_i^t \in \mathbb{R}^{H \times W \times 3} \), where \( t \in \{1, \dots, \mathcal{T}_i\} \) and \( \mathcal{T}_i \) represents the total number of video frames for the \( i^{\text{th}} \) instance. Here, \( H \) and \( W \) denote the height and width of the extracted frames, respectively.
We use a frozen CLIP image encoder \( f_v \) to extract visual features \( \mathbf{E_i^t} \in \mathbb{R}^{\mathcal{N} \times \mathcal{C}} \) from \( V_i^t \), where \( \mathcal{N} = \frac{H}{p} \times \frac{W}{p} \), with \( p \) representing the patch size, and \( \mathcal{C} \) being the embedding dimension.
Additionally, We extract the text-guided query \( \mathbf{Q}_i \in \mathbb{R}^{\mathcal{M} \times \mathcal{C}} \) using a pre-trained RoBERTa \cite{liu1907roberta} model for the question $Q_i$, where \( \mathcal{M} \) denotes the number of queries.
The visual features \( \mathbf{E}_i^t \) and the text-guided query features \( \mathbf{Q}_i \) are then passed through learnable projection \(\phi_e\) and \(\phi_q\) layers to obtain the projected features \( \mathbf{\Tilde{E}}_i^t \) and \( \mathbf{\Tilde{Q}}_i \), respectively. These projected features are subsequently fed into the cross-attention module defined in Eq. \ref{eq:1}, where a cross-attention score between the question and the $t$-th frame, \( s_t \), is computed as follows,
\begin{equation}
    \mathbf{\Tilde{E}}_i^t = \phi_e{(\mathbf{E_i^t})}, \mathbf{\Tilde{Q}}_i = \phi_q{(\mathbf{Q_i})}
\end{equation}
\begin{equation}
    s_t = \operatorname{Mean}\left(\operatorname{Softmax}\left(\mathbf{\Tilde{E}}_i^t \cdot \mathbf{\Tilde{Q}}_i^{\top}\right) \cdot \mathbf{\Tilde{Q}}_i\right)
    \label{eq:1}
\end{equation}

Finally, we select the frame \( V_i^t \) based on the cross-attention score \( s_t \), provided it surpasses a predefined threshold \( \beta \). The subset of selected frames from \( V_i \) is denoted as \( \mathcal{V}_i \). These selected frames are then processed further to construct spatial object graphs for each selected frame.

\subsection{Obtaining graph-based frame representation}

We utilize the MiniGPT-4 architecture to construct question-aware object graphs from the selected frames in \( \mathcal{V}_i \), contrasting with traditional models that perform object detection across all frames without considering the question context. MiniGPT-4's efficiency lies in requiring only a linear layer to align visual features with the Vicuna model\cite{vicuna}. Additionally, we redefine the object detection task in LGQAVE as a visual grounding task, using the frames from \( \mathcal{V}_i \) and the question \( Q_i \), where MiniGPT-4 excels.

For each selected frame \( V_i^{t'} \in \mathcal{V}_i \), we also include the two preceding and two subsequent frames: \( V_i^{t'-2}, V_i^{t'-1} \) through \( V_i^{t'+1}, V_i^{t'+2} \)—to ensure temporal continuity and minimize the risk of missing critical sequential information, which we fixed through empirical validation. These frames, along with the question prompt \(\mathbf{{Q}_i}\) from the frozen RoBERTa model, are fed into MiniGPT-4, which processes them to generate \( m \) bounding boxes \( \mathcal{B}_{i}^{t'} \) around the objects pertinent to the question in $V_i^{t'}$. Four coordinates define each bounding box, and the total number of bounding boxes per frame is limited to \( m \leq 10 \).

We enhance the graph representation methodology by utilizing detected objects, advancing beyond the approach in \cite{CoVGT}. For each highlighted object instance in a video frame \( V_i^{t'} \), we extract Region of Interest (RoI)-aligned features as object appearance representations \( F_o^{t'} \) which also contains spatial locations \( F_s^{t'} \) of the respective objects. Additionally, we capture a frame-level feature \( F_I^{t'} \) to augment the graph representations derived from the local objects. We aim to construct a frame-specific spatial graph using \( F_u^{t'} = F_o^{t'} \cup F_I^{t'} \).

While our methodology is inspired by \cite{CoVGT}, it differs significantly in its execution. Unlike \cite{CoVGT}, which assumes static object groups within a video clip and employs a fixed linking score based on appearance and spatial location  \( F_s^{t'} \), our approach with MiniGPT-4 dynamically tracks objects across the video sequence. This dynamic tracking enhances the robustness and adaptability of our model, particularly improving its generalizability to longer video sequences.

\noindent \textbf{Graph construction for $V_i^{t'}$}: We propose to consider the bounding boxes from $\mathcal{B}_i^{t'}$ and the entire frame $V_i^{t'}$ as constituting the $m+1$ nodes in the frame-specific graph $\mathcal{G}_i^{t'}(A^{t'}, R^{t'})$, with $A^{t'}$ denoting the node-set, and we put up an edge between two bounding boxes, and the edge weights are defined as follows,
\begin{equation}
    R^{t'} = \operatorname{Softmax}\left( \phi_{{k}}(F^{t'}_{u}) \phi_{{v}}(F^{t'}_{u})^\top \right)
\end{equation}

Here, \( \phi_{{k}} \) and \( \phi_{{v}} \) denote linear transformations and the transpose operation is denoted by \( (\cdot)^\top \). The obtained $\mathcal{G}_i\ = \left\{ \mathcal{G}_i^{1}, \mathcal{G}_i^{2}\,\ldots \right\}$ which contains the object representation and also the spatial representations \( F_s^{t'} \) are passed to the Q-DGT module for video feature extraction.



\subsection{Question-aware dynamic graph transformer} 
Following the methodologies proposed in \cite{CoVGT}, we utilize DGT to capture the complex dynamics of objects from the obtained graphs. However, different from \cite{CoVGT}, to enhance the relevance of the object dynamics to the specific questions posed, we condition the DGT on the question (Q-DGT), focusing the analysis only on the objects that are essential for answering the questions. This conditional approach ensures that our model's attention is selectively tuned to the pertinent elements of the video content. Furthermore, to enhance the contextual relevance of the visual information extracted by the Q-DGT module, our approach goes beyond the typical refinement processes described in \cite{CoVGT}, which focuses solely on global representations. We extend refinement to both global and local representations, thereby improving the accuracy and contextual depth of the answer prediction.

In Q-DGT, the question embedding \(\mathbf{\Tilde{Q}}_i\) is intentionally masked to control the influence of the question representation on the model. This masking helps isolate specific features, mitigating the risk of overfitting by finely tuning the interaction between the question representation and the object dynamics captured by the DGT. Such an approach ensures that only the most relevant dynamics are emphasized, enhancing the model's accuracy and generalizability. \\
\begin{equation*}
     \hat{\mathbf{Q}} = \mathbf{M} \odot \mathbf{\Tilde{Q}}_i
\end{equation*}
Here, \(\mathbf{M}\) is binary mask vector, \(\odot\) denotes the element-wise (Hadamard) product.

Q-DGT integrates both a temporal and a spatial graph transformer unit to process the input visual graphs within \(\mathcal{G}_i\). \( F_s^{t'} \) is the input to the spatial unit that models the spatial relationships within each frame, and \( F_u^{t'} \) is the input to the temporal unit that captures relationships across different frames. These units are specifically designed to handle the different dimensions of data - temporal changes over time and spatial relationships within frames. The output of the Q-DGT is a local representation $\mathcal{F}_{local}^{t'}$ corresponding to the $t'$-th frame, which is obtained by non-linearly transforming the embeddings from the frame-specific graph through a trainable projection layer with parameters ${\phi}_{local}$. This projection layer is crucial as it ensures that vital information pertinent to the video is preserved and not lost in transformation processes. The local representation formula is: 
\begin{equation}\centering
\mathcal{F}_{{local}_i}^{t'} = {\phi}_{local}(\text{Q-DGT}(\mathcal{G}^{t'}_i,\hat{\mathbf{Q}}))
\end{equation}

Additionally, a global representation $\mathcal{F}_{global}$ is derived by aggregating all the spatial and temporal representations through a global transformer, similar to the approach in \cite{CoVGT}. This global transformer incorporates learnable sinusoidal temporal position embeddings to model the sequence of events within the video effectively. The outputs of this transformer are then mean-pooled to produce a comprehensive global representation $\mathcal{F}_{global}$ of the entire video, which encapsulates both the spatial and temporal dynamics across all processed frames. MHSA stands for Multihead Self Attention, and MPool represents the Maxpooling operation. 
\begin{equation}\centering
\mathcal{F}_{{global}_i} = \text{MPool}(\text{MHSA}(\text{Q-DGT}(\mathcal{G}_i, \hat{\mathbf{Q}})))
\end{equation}
For further details regarding DGT, refer \cite{CoVGT}. 


\noindent \textbf{Interaction of the question and graph features in Q-DGT:}
To integrate textual context effectively, we employ a RoBERTa language model to process the question \({Q}\) and project the token outputs into a textual information space \( Z_{\hat{\mathbf{Q}}} \) using a linear transformation:
\[
Z_{\hat{\mathbf{Q}}} = \phi_{{\hat{{Q}}}}(\hat{\mathbf{Q}}) = \{z_{\hat{\mathbf{Q}}}^h\}_{h=1}^{\mathcal{H}},
\]
where \(\phi_{{\hat{{Q}}}}\) is a projection matrix in \(\mathbb{R}^{768 \times d}\), \(\mathcal{H}\) denotes the number of tokens in \(\hat{\mathbf{Q}}\), and \(z_{\hat{\mathbf{Q}}}^h\) represents the embedded representation of the \(h^{\text{th}}\) token. The encoded tokens include those representing the words of an open-ended question \(Q\) or QA pairs in a multiple-choice format.

Within the DGT framework, the cross-modal encoder \( Q\text{-}DGT_{cm} \) processes the textual embeddings \( Z_{\hat{\mathbf{Q}}} \) together with the visual embeddings \( \mathbf{\Tilde{E}^{t'}_{i}} \) corresponding to the frame \( \hat{V}^{t'}_i \) for the \(i^{th}\) instance. This integration facilitates a nuanced refinement of both local and global video representations:
\begin{equation}
\mathcal{F}_{{local}} = Q\text{-}DGT_{cm}(\mathcal{F}_{{local}}, Z_{\hat{\mathbf{Q}}}) = \mathcal{F}_{{local}} + \sum_{h=1}^{\mathcal{H}} \alpha^{1}_{h} z_{\hat{\mathbf{Q}}}^{h},
\end{equation}
\begin{equation}
\mathcal{F}_{{global}} = Q\text{-}DGT_{cm}(\mathcal{F}_{{global}}, Z_{\hat{\mathbf{Q}}}) = \mathcal{F}_{{global}} + \sum_{h=1}^{\mathcal{H}} \alpha^{2}_{h} z_{\hat{\mathbf{Q}}}^{h},
\end{equation}
where \(\alpha^{1}\) and \(\alpha^{2}\) are attention weights. These weights are calculated by applying a sigmoid function \( \sigma \) to the transpose dot product between \(\mathcal{F}_{{local}}\) or \(\mathcal{F}_{{global}}\) and \( Z_{\hat{\mathbf{Q}}} \), emphasizing the dynamic and context-sensitive interactions between the modalities.
\(\alpha^{1}\) = \(\sigma(\mathcal{F}_{{local}})\) \(\odot\) \( Z_{\hat{\mathbf{Q}}} \), 
\(\alpha^{2}\) = \(\sigma(\mathcal{F}_{{global}})\) \(\odot\) \( Z_{\hat{\mathbf{Q}}} \).

\subsection{Obtaining the final video features}
Our method acknowledges the dynamic relevance of global image context and local properties based on the question posed. To adeptly handle this variability, we introduce an adaptive mechanism that updates the global embedding through a cross-attention process with the local embeddings obtained from the Q-DGT module.

The final representation of the answer leverages cross-attention between the local representations \( \{\mathcal{F}_{local}^{t'}\} \) and the global representation \( \mathcal{F}_{global} \). Directly merging these representations often leads to redundancy due to overlapping information. To address this issue, our attention mechanism is designed such that \( \mathcal{F}_{global} \) serves as the query, while \( \{\mathcal{F}_{local}^{t'}\} \) function as both keys and values. This structure allows the model to dynamically emphasize the most relevant details from the local context when updating the global representation. This results in a more discriminative and contextually refined final representation.
\begin{equation}
\mathcal{F}_{final} = (1-\gamma) \mathcal{F}_{global} + \gamma \text{Cross-Att}(\mathcal{F}_{global}, \{\mathcal{F}_{local}^{t'}\}),
\end{equation}

$\gamma$ is a weighting constant within the range $[0,1]$.
\( \mathcal{F}_{final} \) embodies a comprehensive, question-aware representation of the video. It seamlessly integrates the broad contextual overview provided by the global features with the detailed insights offered by the local features.

\subsection{Answer generation}
In our framework, we employ distinct strategies for answering objective and subjective questions, leveraging the synthesized representation \( \mathcal{F}_{final} \).

For objective questions, the answer prediction \(\hat{A}\) is determined by calculating the similarity scores between \( \mathcal{F}_{final} \) and a set of pre-encoded answer representations \( F_A \). Here, \( A = \{ A_{l} \}_{l=1}^{|A|} \), where \( |A| \) represents the number of answer options, and \( \mathbf{A}_{l} \) denotes the RoBERTa-encoded representation of each option \( l \). The prediction is made by identifying the option associated with the highest similarity score:
\begin{equation}\centering
\hat{A} = \arg\max \left( (\mathcal{F}_{final})^\top \mathbf{A} \right)
\end{equation}

We adopt a methodology for subjective questions that enables a video-absent QA scenario, as delineated in prior works \cite{CoVGT}. The answer is inferred by evaluating the similarities not only between \( \mathcal{F}_{final} \) and \( A \) but also between the question representation \( \mathbf{\Tilde{Q}} \) and \( \mathbf{A} \). The final prediction \(\hat{A}\) is obtained by taking an element-wise product of these similarity matrices, thereby ensuring that the decision robustly integrates cues from both the video and the question:
\begin{equation}\centering
\hat{A}\ = \arg\max \left( (\mathcal{F}_{final})^\top \mathbf{A} \odot (\mathbf{\Tilde{Q}})^\top \mathbf{A} \right)
\end{equation}

\subsection{Loss objectives}

\noindent \textbf{Loss function for multi-choice QA:}
We employ a composite loss function in multi-choice question answering, where answers are selected from given options. The component \( L_{\text{vqa}} \) accounts for the interaction between the video, the question, and the multiple-choice options, while \( L_{\text{vq}} \) pertains solely to the video and the question:
\begin{equation}
L = L_{\text{vqa}}(\mathcal{F}_{final}, \mathbf{\Tilde{Q}} \otimes \mathbf{A}^{+}, \mathbf{\Tilde{Q}} \otimes \mathbf{A}^{-}) + \lambda L_{\text{vq}}(\mathcal{F}_{final}, \mathbf{{Q}}^{+}, \mathbf{{Q}}^{-})
\end{equation}
\( A^{+} \) and \( A^{-} \) represent the correct and incorrect answer options, respectively. Similarly, \( Q^{+} \) and \( Q^{-} \) denote the positive and negative questions associated with a video. The balancing parameter is represented by \( \lambda \), and the symbol \( \otimes \) indicates a concatenation operation.

\noindent \textbf{Loss function for open-ended QA:}
For open-ended QA, where the answer \( a \) is not constrained to predefined options, the loss formulation needs to adapt to the broader scope of potential answers:
\begin{equation}
L = L_{\text{vqa}}(\mathcal{F}_{final} \otimes \mathbf{\Tilde{Q}}, \mathbf{A}^{+}, \mathbf{A}^{-}) + \lambda L_{\text{vq}}(\mathcal{F}_{final}, \mathbf{\Tilde{Q}}^{+}, \mathbf{\Tilde{Q}}^{-})
\end{equation}


\section{Experimental Evaluations}
\noindent \textbf{Datasets$^{\dagger}$: :}
We conduct experiments across various datasets to evaluate different aspects of video understanding. The datasets include NExT-QA \cite{nextqa}, STAR-QA \cite{starqa}, and Causal-VidQA \cite{casualqa}, which are designed to address complex temporal and causal relationships as well as commonsense reasoning within videos, with a particular focus on temporal dynamics. Additionally, we utilize TGIF FrameQA \cite{tgifqa}, MSRVTT-QA \cite{msrvttqa}, and ActivityNetQA \cite{ActivityNetQA}, which concentrate on the recognition of video objects, their attributes, actions, and activities, emphasizing static frame analysis.

\footnotetext[1]{$^{\dagger}$More about datasets and implementation details in \textcolor{blue}{supplementary material.}}
\subsection{Main results}

\begin{table*}[t]
\centering
\caption{Accuracy (\%) comparison on NExT-QA \cite{nextqa}, TGIF-FrameQA\cite{tgifqa}, MSRVTT-QA\cite{msrvttqa} and ActivityNet-QA\cite{ActivityNetQA}. Acc@C, T, and D denote accuracy for Causal, Temporal, and Descriptive questions. The \textbf{best} and \underline{2nd best} results are highlighted in bold and underlined, respectively.}
\label{tab:results}
\scalebox{0.6}{
\begin{tabular}{l|c|c|c|c|c|c|c|c|c|c}
\hline
 \multirow{2}{*}{\textbf{Methods}} & \multirow{2}{*}{\textbf{Text}} & \multicolumn{2}{c|}{\textbf{NExT-QA Val}} & \multicolumn{2}{c|}{\textbf{NExT-QA Test}} & \multirow{2}{*}{\textbf{TGIF-FrameQA}} & \multirow{2}{*}{\textbf{MSRVTT-QA}} & \multirow{2}{*}{\textbf{ActivityNet-QA}} & \multirow{2}{*}{\textbf{Star-QA}} & \multirow{2}{*}{\textbf{Causal-VidQA}} \\
\cline{3-6}
 &  & Acc@C & Acc@T & Acc@D  & Acc@All & & & & &\\
\hline
\cellcolor{red!20}VQA-T \cite{vqa-t} & DistilBERT & 41.66 & 44.11 & 59.97 & 45.30 &25.30 &40.40 &15.70 &29.61 &40.32 \\ 
\cellcolor{red!20}HGA \cite{hga} & BERT & 46.26 & 50.74 & 59.33 & 51.02 &20.70 &31.53 &14.82 &32.27 &44.82\\ 
\cellcolor{red!20}HQGA \cite{hqga} & BERT & 48.48 & 51.24 & 61.65 & 51.34 &25.40 &33.80 &17.51 &35.83 &47.36 \\ 
\cellcolor{red!20}ATP \cite{buch2022} & BERT & 51.57 & 52.00 & 66.80 & 53.18 &26.33 &31.76 &16.47 &39.27 &50.14\\
\cellcolor{red!20}VGT \cite{VGT}& BERT & 52.28 & 55.09 & 61.94 & 53.68 &61.60 &39.70 &20.40 &42.43 &53.20\\ 
\cellcolor{red!20}VGT (PT) \cite{VGT}& BERT & 53.43 & 56.39 & 59.64 & 55.70 &61.70 &3.70 &19.70 &44.32 &54.35\\
\cellcolor{red!20}CoVGT\cite{CoVGT} & RoBERTa & 58.53 & 57.48 & 63.82 & 57.40 &61.60 &38.30 &24.50 &46.20&60.80\\ 
\cellcolor{blue!20}VideoChat\cite{videochat} & - &62.30  &59.36  &64.22  &56.27  &34.40 &45.00 &26.50 &49.35 &66.64\\ 
\cellcolor{blue!20}VideoLlama\cite{videollama} & - &61.53  &61.25  &66.35  &58.41  &- &29.60 &12.40 &53.47 &68.35\\ 
\cellcolor{blue!20}VideoLlava\cite{videollava} & - &\underline{63.70}  &\underline{63.45}  &\underline{69.10}  &\underline{60.08}  &\underline{70.00} &\underline{59.20} &\textbf{45.30} &\textbf{62.25} &\underline{70.31}\\ \hline
\textbf{LGQAVE} & RoBERTa & \textbf{68.69} & \textbf{68.00} & \textbf{74.88} & \textbf{66.69} &\textbf{72.40} &\textbf{63.43} &\underline{44.81} &\underline{61.48} &\textbf{73.59} \\ \hline
\end{tabular}
}
\vspace{-0.5cm}
\end{table*}

 We compare LGQAVE with several relevant and recent methods from the literature in Table \ref{tab:results} on all the datasets mentioned above. LGQAVE significantly surpasses the previous SOTAs on all tasks defined in previously mentioned datasets, improving the accuracy on an average by 9.29\% vs. non-LLM methods like CoVGT \cite{CoVGT} and 6.61\% vs. LLM models like VideoLlava \cite{videollava}, respectively. Compared to other methods, we paid more attention to the video content related to the question instead of taking all the video content features and giving answers based on such features. A sampling of frames gained a lot of popularity recently and fine grained frame selection method as shown in \cite{Nuthalapati_2023_ICCV} works much better than other sampling methods. \\
 Methods like VideoLlava and VideoChat, which do not use graph-based approaches, generalize well for tasks like video summarization and captioning but struggle with reasoning and frame-level questions, particularly in longer videos. In various datasets, we have seen that existing methods struggle to answer reasoning questions on videos with more than 600 lengths, primarily when it is based on a few frames in the video. VideoGPT addresses this by using frame-level captions, but it requires test videos to match the distribution of training videos. Our approach leverages graphs and the LLM model miniGPT, focusing on video understanding without relying on captions, particularly excelling in object recognition. Our methods takes approximately 289GFlops during training and 138GFlops during testing.

\noindent\textbf{Our frame selection vs coarse frame selection \cite{ge2022bridgeformer}:} Compared with the coarse frame selection process, which generally employs BridgeFormer\cite{ge2022bridgeformer}, our frame selection (FFS) is better in picking up the correct frames related to the question. An average increase of 4.23\% is observed with this method alone, as shown in Table \ref{tab:frameselection}. BridgeFormer \cite{ge2022bridgeformer} concentrates on nouns and verbs from the question, removing the remaining phrase of the question. In comparison, the fine-grained frame selection process takes all the parts of speech in the question into context, which makes it robust in selecting appropriate frames related to the question. 

\noindent\textbf{LGQAVE vs. graph methods + sampling:} To show the supremacy of our model, we conducted thorough experiments by including frame sampling methods with the existing graph methods as shown in Table \ref{tab:frameselection}.  Improved the existing HQGA \cite{hqga}, existing CoVGT \cite{CoVGT} by adding the frame sampling modules in their architectures. We made two versions of these architectures, one by adding a coarse frame selection method \cite{ge2022bridgeformer} and another by adding a finer frame selection method \cite{Nuthalapati_2023_ICCV}. LGQAVE architecture works better than any other graph video question-answering model even after the frame selection process, which shows that making graphs using objects specific to the question and their local and global representations gives an advantage to our model for a better understanding of the question and answering it. 
\begin{figure*}[h]
    \centering   
    \scalebox{0.8}{\includegraphics[width=0.33\linewidth]{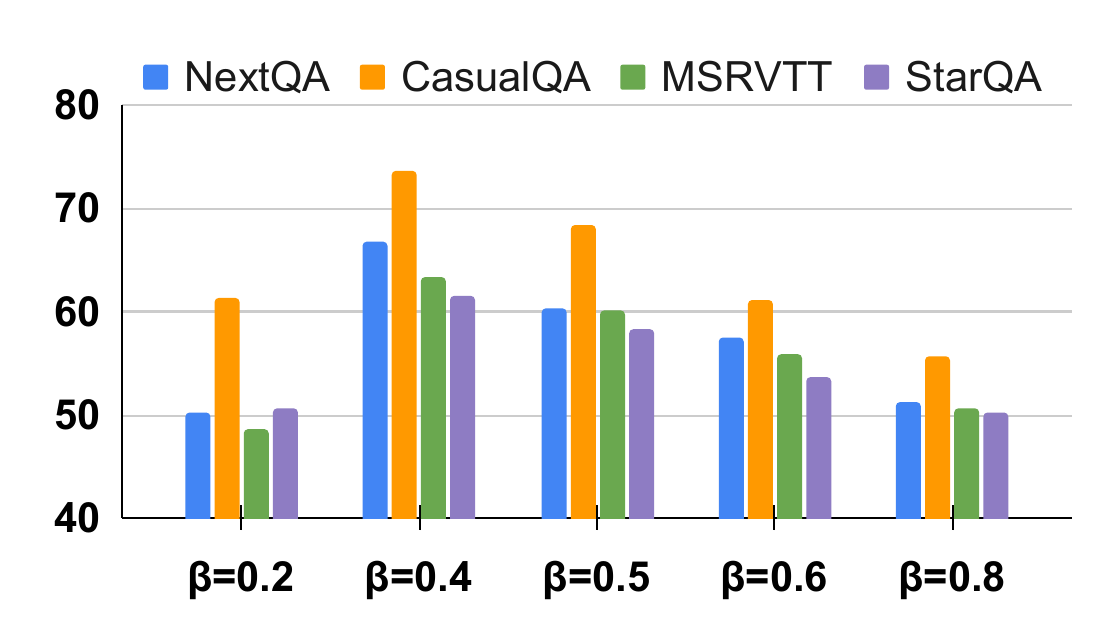}
    \includegraphics[width=0.33\linewidth]{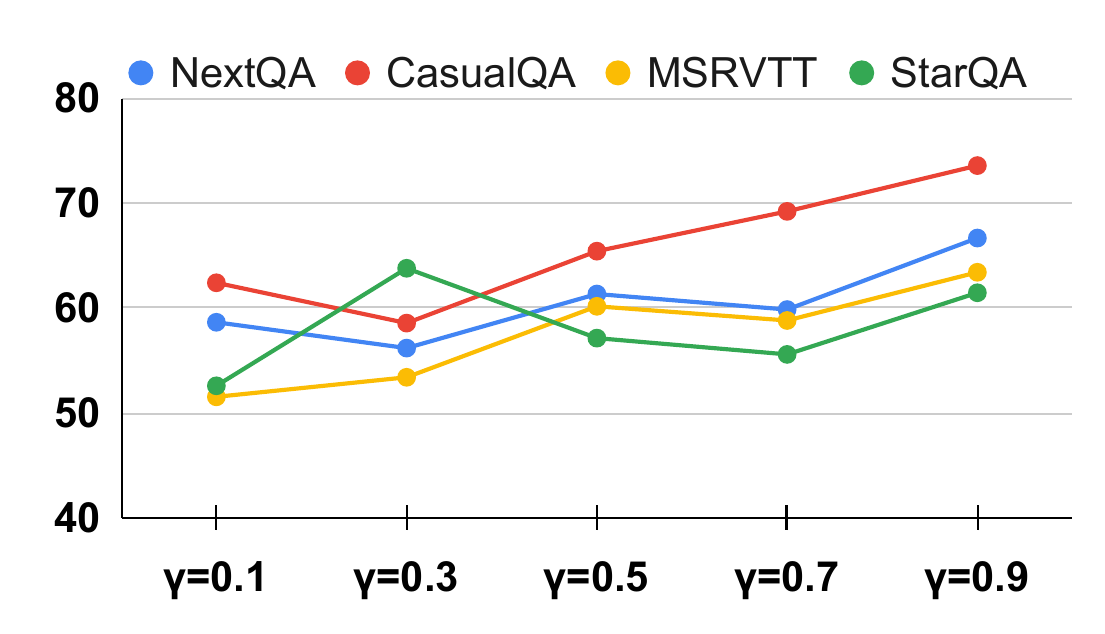}
    \includegraphics[width=0.33\linewidth]{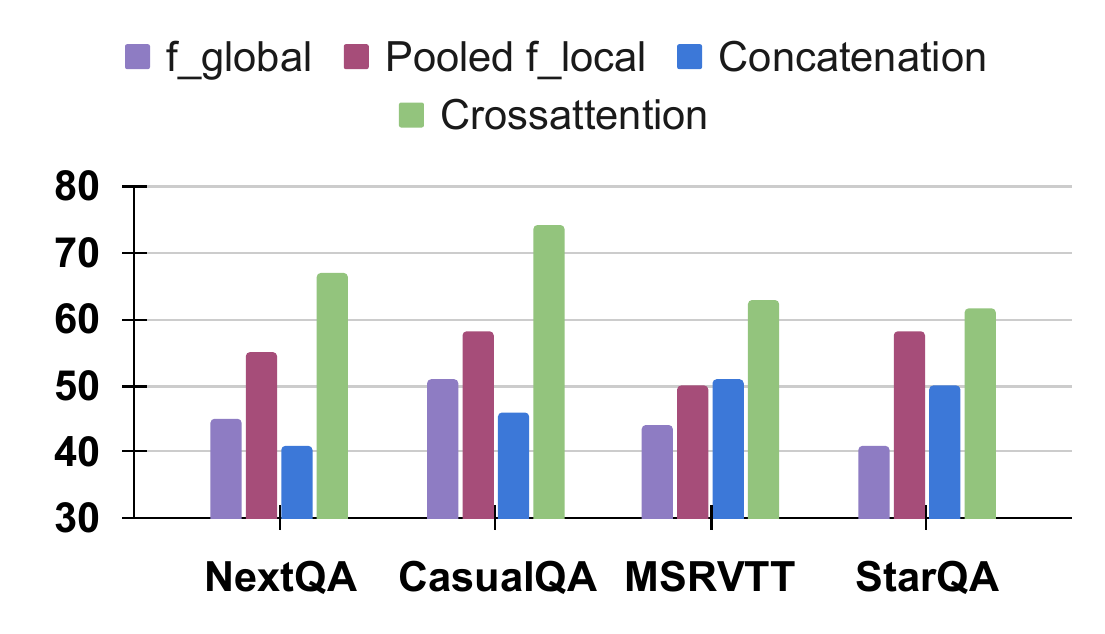}}
    \caption{Performance of LGQAVE with change in \(\beta\) and \(\gamma\) parameters on various datasets are shown in the first two plots. Performance of LGQAVE with usage of different combinations of \(\mathcal{F}_{{local}}\) and \(\mathcal{F}_{{global}}.\) is shown at the end. }
    \label{fig:comparisons}
    \vspace{-0.7cm}
\end{figure*}
\begin{table}[h!]
\centering
\vspace{-0.5cm}
\caption{Detailed comparison between LGQAVE and other SOTA methods for frame sampling. CFS\cite{ge2022bridgeformer}: coarse frame selection\cite{Nuthalapati_2023_ICCV}. FFS: fine frame selection.}
\label{tab:frameselection}
\scalebox{0.8}{
\begin{tabular}{l|c|c|c|c}
\hline
\multirow{2}{*}{Models} & \multicolumn{4}{c}{NExT-QA Val} \\ \cline{2-5}
   & Acc@C & Acc@T & Acc@D & Acc@All \\ \hline
HQGA+CFS  & 50.66 & 54.11 & 59.97 & 48.30 \\ 
HQGA+FFS   & 52.27 & 53.29 & 62.17 & 49.40 \\
CoVGT+CFS  & 61.62 & 59.08 & 66.42 & 59.02 \\ 
CoVGT+FFS   & 64.31 & 62.27 & 69.50 & 61.47 \\ \hline
LGQAVE (RoBERTa)  & \textbf{68.69} & \textbf{68.00} & \textbf{74.88} & \textbf{66.69} \\ \hline
\end{tabular}}
\vspace{-0.5cm}
\end{table}


\noindent\textbf{Comparison of other graph-based methods with frame selection process:} Table \ref{tab:frameselection} compares our LGQAVE model to other state-of-the-art graph-based methods on the NExT-QA validation set. The HQGA models, employing either Coarse Frame Selection (CFS) or Fine Frame Selection (FFS), show limited performance, with \texttt{Acc@All} at only 49.40\%. CFS, which selects a broad range of frames for a general video overview, often includes irrelevant frames and misses finer details crucial for precise answers, resulting in lower accuracy for HQGA+CFS and CoVGT+CFS. Conversely, FFS targets the most relevant frames, focusing on specific objects or actions linked to the questions, thus improving accuracy. This method filters out extraneous content, concentrating on critical frames and leading to higher \texttt{Acc@All} scores for HQGA+FFS and CoVGT+FFS.

Our LGQAVE model, enhanced with RoBERTa, significantly outperforms both methods with an \texttt{Acc@All} of 66.69\%, demonstrating the effectiveness of integrating local and global visual features with advanced textual encoding for more accurate, context-aware video question answering. The results underscore the advantages of LGQAVE, especially when combined with sophisticated language models, in leveraging both detailed visual representations and broader scene context.
\begin{figure}[h]
    \centering
    
       \hspace{-9.5mm} \includegraphics[width=0.6\linewidth]{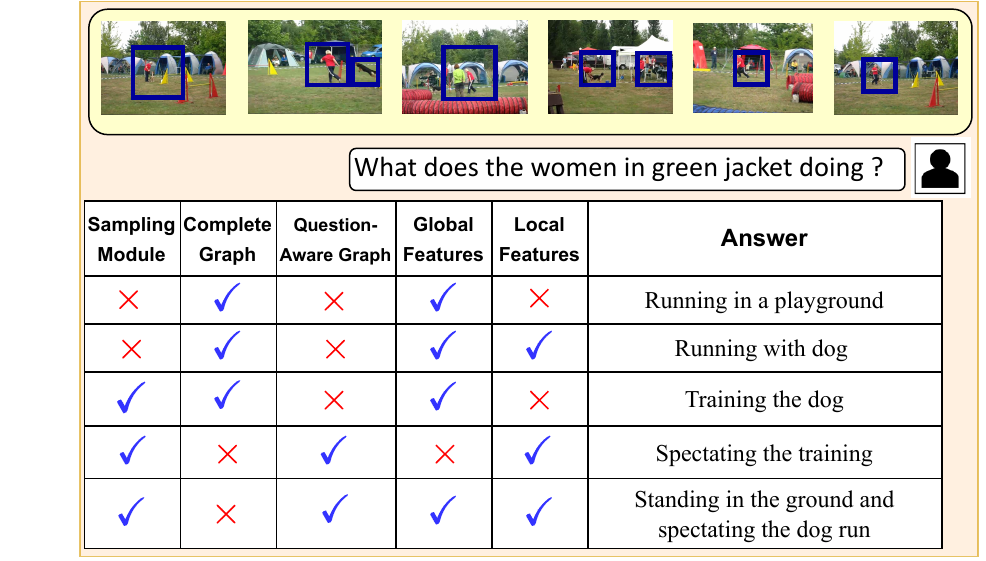}
    
    \caption{Qualitative answers$^{\dagger}$ by LGQAVE model for various ablation configurations on a video from the NextQA dataset.}
    \vspace{-0.75cm}
    \label{fig:visab}
\end{figure}
\section{Ablation analysis}

To better understand the contribution of each component
in the proposed model LGQAVE, we conducted ablation over the
various components, shown in Table \ref{tab:configuration}. The inclusion of a sampling strategy markedly enhances our model's performance. Without sampling (Configuration C-1), the model depends solely on global representations, which limits its focus on pertinent frames and leads to reduced accuracy, notably in \texttt{Acc@All}. Introducing sampling in Configuration C-2 improves focus on relevant frames, resulting in significant performance gains across all metrics, particularly in \texttt{Acc@C} (\(+3.36\%\)) and \texttt{Acc@T} (\(+3.15\%\)), by filtering out extraneous information.
\begin{table}[!htbp]
\vspace{-2mm}
\caption{Ablation analysis of the proposed model components of \textsc{LGQAVE} on the NExT-QA dataset.}
\label{tab:configuration}
 \vspace{-5mm}
\begin{center}
\scalebox{0.5}{
\begin{tabular}{l|c|c|c|c|cccc}
\hline
\multirow{2}{*}{\textbf{Conf.}} & \multirow{2}{*}{\thead{Sampling}}  & \multirow{2}{*}{\thead{miniGPT}} & \multirow{2}{*}{local Repr.} & \multirow{2}{*}{Global Repr.}
& \multicolumn{4}{c}{\textbf{NExT-QA}} \\ 
\cline{6-9} 
 &  &  &  &  & \texttt{Acc@C} & \texttt{Acc@T} & \texttt{Acc@D} & \texttt{Acc@All} \\ 
\hline
C-1 & $\times$ & $\times$ & $\times$ & $\checkmark$ & 58.53 & 57.48 & 63.82 & 57.40 \\
C-2 & $\checkmark$ & $\times$ & $\times$ & $\checkmark$ & 61.89 & 60.63 & 65.37 & 61.85 \\
C-3 & $\checkmark$ & $\checkmark$ & $\times$ & $\checkmark$ & 65.42 & 64.79 & 71.53 & 64.76 \\
C-4 &  $\checkmark$ & $\times$ & $\checkmark$ & $\checkmark$ &59.26  &56.18  &57.46  & 58.13 \\
C-5 & $\checkmark$ & $\checkmark$ & $\checkmark$ & $\checkmark$ & \textbf{68.69} & \textbf{68.00} & \textbf{74.88} & \textbf{66.69} \\ 
\hline
\end{tabular}}
\end{center}
\vspace{-8mm}
\end{table}

The integration of miniGPT in C-3, combined with sampling but excluding local representations, significantly enhances accuracy, particularly in \texttt{Acc@D} (\(+6.16\%\)), suggesting that miniGPT enriches the model’s contextual understanding and response accuracy. In contrast, using graphs with all objects in a frame and employing both local and global representations leads to a severe drop in accuracy, as observed in  C-4. In C-5, we leverage local and global representations by cross-attention to balance detailed object-level insights and broader scene context, resulting in the highest accuracy across all metrics. This approach outperforms models that rely solely on global features by \(+9.29\%\) in \texttt{Acc@All}. 

In Figure \ref{fig:comparisons}, we analyze the impact of varying the parameter 
\(\beta\) and \(\gamma\) across four datasets: NextQA, CasualQA, MSRVTT, and StarQA. The study suggests that the optimal \(\beta\) and \(\gamma\) are 0.4 and 0.9, respectively, where the highest performance is observed, with performance declining at higher or lower values. Also, we show that our cross-attention of \(\mathcal{F}_{{local}}\) and \(\mathcal{F}_{{global}}\) gives better accuracy than using \(\mathcal{F}_{{global}}\) or pooled \(\mathcal{F}_{{local}}\) or concatenating them.  

Fig \ref{fig:visab} highlights the impact of various model configurations on the task of answering video questions. The sampling module, made of graphs and global and local features, was individually assessed for their contribution to the model's overall performance. By isolating each module, the study reveals that using the question-aware object interaction graphs in combination with Local Features significantly improves the accuracy of the model’s predictions. For instance, it enables the model to generate more specific and contextually appropriate answers, such as distinguishing between actions like "Training the dog" and "Spectating the dog run." This suggests that incorporating both question-awareness and fine-grained local features plays a crucial role in understanding video content.
\vspace{-0.3cm}

\footnotetext[2]{$^{\dagger}$More qualitative results in \textcolor{blue}{supplementary material}.}

\section{Takeaways}
We present LGQAVE, a novel framework that addresses limitations in existing VideoQA approaches by enhancing multi-modal integration and focusing on semantic visual concepts relevant to the questions. Using cross-attention, LGQAVE identifies the most pertinent video frames for each query, surpassing traditional frame sampling techniques. Our approach generates precise video representations by capturing object dynamics through spatial graphs and grounding them in question semantics via the MiniGPT model. Q-DGT refines these representations, ensuring global and local video content is optimally encoded. An additional cross-attention module synthesizes final video embeddings conditioned on the questions, leading to more accurate answer generation by the language model.
Extensive evaluations across benchmarks show that LGQAVE significantly improves accuracy in multi-choice and open-ended VideoQA tasks, suggesting future opportunities to leverage advanced graph-based and attention mechanisms for multi-modal integration.
\vspace{-0.3cm}
\section*{Acknowledgments}
This work was conducted in collaboration with Fractal AI Research Team, who also provided the financial support necessary for this research.We gratefully  acknowledge their contribution and support.

{\small
\bibliographystyle{ieee_fullname}
\bibliography{egbib}
}

\end{document}


\title{Supplementary: Foundation Models and Adaptive Feature Selection: A Synergistic Approach to Video Question Answering}  
\author{Sai Bhargav Rongali\\
Indian Institute of Technology Bombay, India\\
{\tt\small rongalisaibhargav002@gmail.com}
\and
Mohamad Hassan N C\\
Indian Institute of Technology Bombay, India\\
{\tt\small  mohdhassannc@gmail.com}
\and 
Ankit Jha\\
LNMIIT, Jaipur, India\\
{\tt\small ankitjha16@gmail.com}
\and 
Neha Bhargava\\
Fractal AI Research, India\\
{\tt\small neha.bhargava@fractal.ai}
\and 
Saurabh Prasad\\
Univesity of Houston\\
{\tt\small saurabh.prasad@ieee.org}
\and
Biplab Banerjee\\
Indian Institute of Technology Bombay, India\\
{\tt\small  getbiplab@gmail.com}
}
\maketitle
\thispagestyle{empty}
\appendix

\begin{table*}[ht!]
\centering

\caption{Overview of the datasets used in the experiments.}
\label{tab:datasets_overview}
\scalebox{1}{
\begin{tabular}{l|c|c|c|c}
\hline
\textbf{Dataset} & \textbf{\#Videos/\#QAs} & \textbf{Train} & \textbf{Val} & \textbf{Test} \\ \hline
NExT-QA \cite{nextqa} & 5.4K / 48K & 3.8K / 34K & 0.6K / 5K & 1K / 9K \\ \hline
TGIF-QA \cite{tgifqa}  & 91.8K / 134.7K & 79.2K / 112.6K & - & 12.5K / 22.2K \\ \hline
ActivityNet-QA \cite{ActivityNetQA} &5.8K/58K   &4.64K /46.4K & - &1.16K/11.6K  \\ \hline
STAR-QA \cite{starqa} & 5K / 60K & 3K / 46K & 1K / 7K & 1K / 7K \\ \hline
Causal-VidQA \cite{casualqa} & 26.9K / 161.4K & 18.8K / 112.7K & 2.7K / 16.0K & 5.4K / 32.6K \\ \hline
MSRVTT-QA \cite{msrvttqa} & 10K / 244K & 6.5K / 159K & 0.5K / 12K & 3K / 73K \\ \hline
\end{tabular}}

\end{table*}
\section{Contents of the Supplementary Material}

In this supplementary material, we present the following details:

\begin{itemize} \item Section \ref{sec:imple} provides detailed information on the dataset used in the experiments. We also discuss our implementation setup for the proposed LGQAVE. \item We study the effects of various loss functions used in the design of our proposed LGQAVE in Section \ref{sec:loss}. \item To showcase the performance of our proposed LGQAVE qualitatively, we present visualization results in Section \ref{sec:vis}. \item Finally, in Section \ref{sec:var}, we summarize the list of variables used in the proposal of LGQAVE. \end{itemize}
\section{Dataset and Implementation details}\label{sec:imple}

\subsection{Dataset overview}

Table \ref{tab:datasets_overview} provides an overview of the datasets used in our experiments. These datasets span various domains and question-answering formats, each contributing to the evaluation of different aspects of video question answering (VQA). \textbf{NExT-QA} \cite{nextqa} focuses on causal and temporal reasoning with 5.4K videos and 48K question-answer pairs. \textbf{TGIF-QA} \cite{tgifqa} is divided into three distinct tasks: Repetition Action, State Transition, and Frame QA, each with varying amounts of video and question data. \textbf{STAR-QA} \cite{starqa}, with its 5K videos and 60K questions, emphasizes situated reasoning. \textbf{Causal-VidQA} \cite{casualqa} pushes the boundaries of evidence-based and commonsense reasoning with a large dataset comprising 26.9K videos and 161.4K questions. Lastly, \textbf{MSRVTT-QA} \cite{msrvttqa}, focusing on visual recognition, provides an extensive dataset of 10K videos and 244K question-answer pairs. These datasets cover a broad spectrum of reasoning tasks and question-answering structures, which ensures the robustness and generalizability of our model.

\subsection{Implementation details}
We processed each video by decoding it into frames and selecting $32$ frames per video. These frames were split into 8 clips, each containing 4 frames, as described in \cite{florence}. To extract features, we used the pre-trained CLIP model ViT-B/32 \cite{clip}, setting the embeddings to 100 tokens per frame and padding with empty tokens if needed. Object detection was performed using the MiniGPT model \cite{minigpt4}, which produced up to 10 graphs per clip, with any unused graphs left empty. The graph model has two layers with hidden states of size $512$, and the transformer module uses one layer with 8 self-attention heads. For edge transformations in the Q-DGT, the self-attention heads are reduced to 5. Frames were selected for further processing based on a cross-attention score greater than $0.4$ ($\beta > 0.4$). We used a decay factor of $\gamma = 0.9$ when computing the final feature representation, $\mathcal{F}_{final}$. In multiple-choice QA, wrong options were used as negative samples, while in open-ended QA, negative samples included answers from other questions and difficult negatives from the same category.
The model was regularized with a parameter $\tilde{\lambda} = 1$ and trained using the Adam optimizer \cite{adam}. We started with a learning rate of $5 \times 10^{-5}$, which decreased over time using a cosine annealing schedule. The batch size was set to 64, and the model was trained for up to 30 epochs, depending on the dataset.

\section{Ablation Study on Loss Functions}\label{sec:loss}

\begin{figure}[h]
    \centering
    \includegraphics[width=0.45\textwidth]{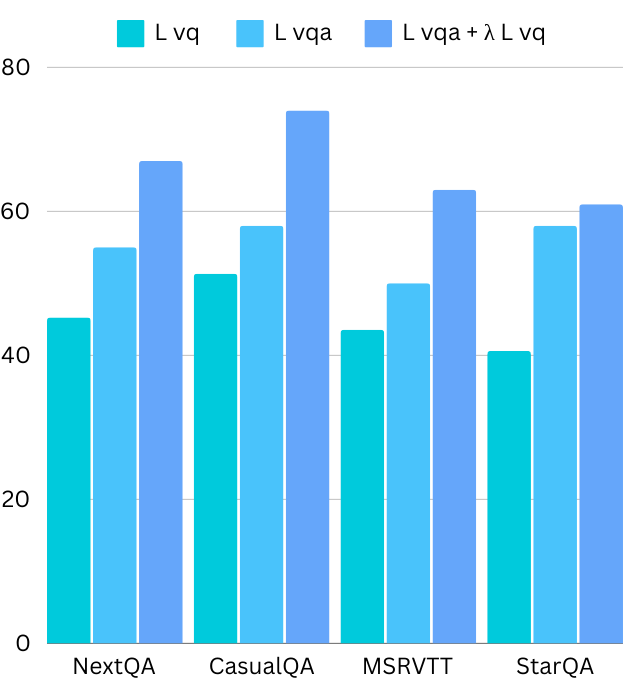}
    \caption{Effect of loss functions on our LGQAVE model.}
    \label{fig:loss_ablation}
\end{figure}

We conducted a detailed ablation study to assess how different components of our composite loss function affect performance in video question answering (VQA) tasks. The plots in Figure~\ref{fig:loss_ablation} show the performance across four major datasets: NextQA, CausalQA, MSRVTT, and StarQA, using different configurations of the loss functions. As described in the main paper, the composite loss function consists of two key components, i.e.,\\
\noindent a) \textbf{\( L_{\text{vq}} \)}, captures the direct interaction between the video and the question.\\
\noindent b) \textbf{\( L_{\text{vqa}} \)}, accounts for the multi-modal interaction between the video, the question, and the multiple-choice options or the answer in an open-ended scenario. To balance the contributions of these components, we introduce a regularization factor \( \lambda \). This leads to the combined loss function:
\begin{equation}\label{eq:total}
L = L_{\text{vqa}} + \lambda L_{\text{vq}}.
\end{equation}

The bar graph shows the performance of the model on the following configurations:
\begin{itemize}
    \item \textbf{\( L_{\text{vq}} \) alone}: Represented in cyan, this configuration uses only the video-question interaction term.
    \item \textbf{\( L_{\text{vqa}} \) alone}: Represented in light blue, it captures the interaction between the video, question, and multiple-choice options.
    \item \textbf{\( L_{\text{vqa}} + \lambda L_{\text{vq}} \)}: Represented in blue, this configuration combines both loss terms with a balancing parameter \( \lambda \).
\end{itemize}

From the results shown in Figure \ref{fig:loss_ablation}, we observe that our proposed LGQAVE performs the worst when trained only with \( L_{\text{vq}} \) loss across all datasets. In contrast, training with \( L_{\text{vqa}} \) loss, which captures the multimodal interaction between the video and the question, shows better performance compared to using only \( L_{\text{vqa}} \) loss. Specifically, the performance improves significantly with \( L_{\text{vqa}} \), especially on the CasualQA and StarQA datasets. This indicates that interactions between the video, question, and multiple-choice options are crucial for accurate question answering. 

We further combine both losses, \( L_{\text{vq}} \) and \( L_{\text{vqa}} \), using the regularization factor \( \lambda \), as described in Equation \ref{eq:total}. This combined loss function achieves the highest accuracy on the NextQA and MSRVTT datasets, highlighting the complementary benefits of both loss terms. Our ablation study demonstrates the effectiveness of using a composite loss function that integrates both video-question interaction and the more complex multimodal interactions. Incorporating \( L_{\text{vqa}} \) significantly enhances model performance, while \( \lambda \) helps balance the contributions of \( L_{\text{vq}} \) and \( L_{\text{vqa}} \), refining the overall results.
\begin{figure*}[ht!]
    \centering
    \includegraphics[width=\textwidth]{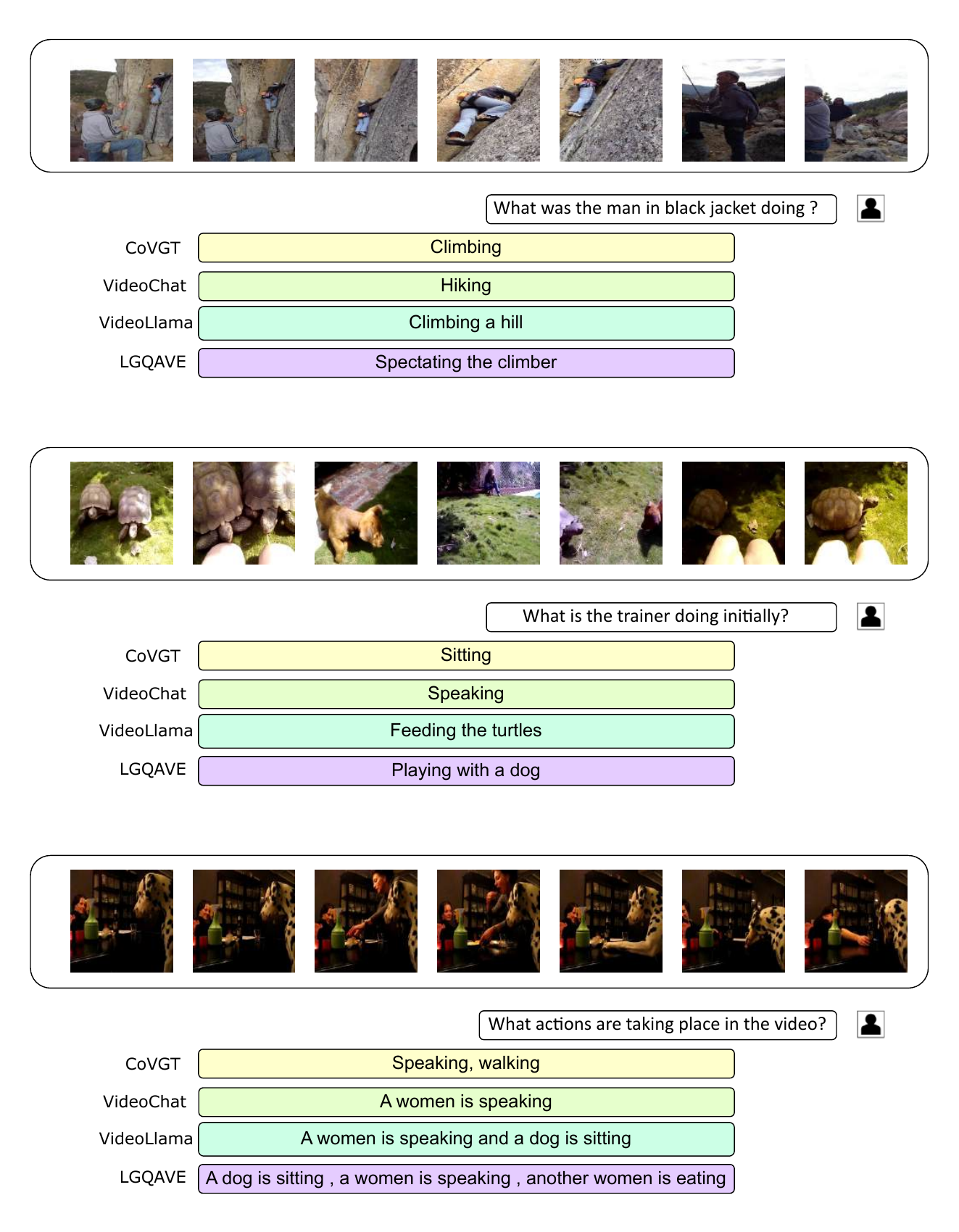}
    \caption{Visual examples of model performance on various video question answering tasks. The model demonstrates its ability to select relevant video segments based on the question and answer accordingly, handling both simple and complex scenarios.}
    \label{fig:visual_results1}
\end{figure*}

\begin{figure*}[ht!]
    \centering
    \includegraphics[width=\textwidth]{images/Visual_ Ablations.pdf}
    \caption{Ablation study of different components showing the strength of our model and the precise answers produced while using all the components. Inaccurate answers when missing various components show their importance. }
    \label{fig:visual_results}
\end{figure*}



\section{Additional visual results}\label{sec:vis}
We showcase the qualitative results of our proposed LGQAVE and compare it with state-of-the-art methods in Figure \ref{fig:visual_results1}. These results demonstrate the model's ability to handle a wide range of video questions, from basic recognition tasks to more advanced reasoning challenges. For instance, as illustrated in Figure \ref{fig:visual_results1}, when asked about a scenario involving a man in a black jacket climbing a hill, the model successfully identified the relevant video segment depicting the action and generated the correct answer. This highlights the capability of model to recognize the video context and the specific action relevant to the question.

In another example, when presented with a question about a trainer’s actions, the model responded with "Feeding the turtles." This response reflects the model’s summarization of the video, likely due to the turtles' activity being more prominent over time. However, in reality, the trainer is seated farther away, feeding a dog. This scenario illustrates the model’s growing proficiency in understanding questions, though it occasionally prioritizes dominant visual cues over subtle actions. Moreover, the model effectively handled complex, multi-object, and multi-action scenarios. For example, in a video featuring a woman speaking, a dog sitting, and another woman eating, the model accurately selected the pertinent segments to answer the question. This reinforces the model’s ability to reason across multiple events and objects, successfully detecting and interpreting simultaneous actions.

These visual results confirm the model’s improved question comprehension and its ability to provide accurate answers based on the relevant temporal segments. They also support the quantitative improvements observed across datasets, demonstrating a robust understanding of video-question interactions. We further discuss the qualitative effects of different modules incorporated in our LGQAVE model in Figure \ref{fig:visual_results}. This includes modules such as the sampling module, complete graph, question-aware graph, global features, and local features. We observe that the sampling module, question-aware graph, and global and local features generate more relevant answers for the video and question queries compared to other settings.

\section{Table of variables}\label{sec:var}
Table \ref{tab:variables} provides a comprehensive list of the key variables used in this paper. The "Description" column outlines the specific roles and applications of each variable within our model, offering clarity on their function and relevance.

\begin{table*}[ht!]

\centering
\label{tab:variables}
\begin{tabular}{|c|p{10cm}|}
\hline
\textbf{Variable} & \textbf{Description} \\
\hline
\( V_i^t \) & The video frame at the \( t^{\text{th}} \) time step for the \( i^{\text{th}} \) instance. Dimensions are \( H \times W \times 3 \), where \( H \) is height and \( W \) is width. \\
\hline
\( \mathcal{T}_i \) & The total number of frames in the \( i^{\text{th}} \) video instance. \\
\hline
\( f_v \) & Frozen CLIP image encoder used to extract visual features from frames. \\
\hline
\( \mathbf{E}_i^t \) & Visual features extracted from \( V_i^t \) using \( f_v \). Dimension is \( \mathbb{R}^{\mathcal{N} \times \mathcal{C}} \), where \( \mathcal{N} \) is the number of patches and \( \mathcal{C} \) is the feature dimension. \\
\hline
\( p \) & The patch size for splitting the image into patches. \\
\hline
\( \mathcal{N} \) & Number of image patches, calculated as \( \mathcal{N} = \frac{H}{p} \times \frac{W}{p} \). \\
\hline
\( \mathcal{C} \) & Embedding dimension of visual features. \\
\hline
\( \mathbf{Q}_i \) & Text-guided query representation for the question \( Q_i \), extracted using a pre-trained RoBERTa model. \\
\hline
\( \mathcal{M} \) & Number of query tokens in the question representation. \\
\hline
\( \mathbf{\Tilde{E}}_i^t \) & Projected visual features after passing \( \mathbf{E}_i^t \) through the learnable projection layer \( \phi_e \). \\
\hline
\( \mathbf{\Tilde{Q}}_i \) & Projected question features after passing \( \mathbf{Q}_i \) through the learnable projection layer \( \phi_q \). \\
\hline
\( s_t \) & Cross-attention score between question \( \mathbf{\Tilde{Q}}_i \) and visual features \( \mathbf{\Tilde{E}}_i^t \) of frame \( V_i^t \), used to select relevant frames. \\
\hline
\( \beta \) & Predefined threshold value for frame selection based on \( s_t \). \\
\hline
\( \mathcal{V}_i \) & Set of selected frames from the \( i^{\text{th}} \) video, based on the cross-attention score \( s_t \). \\
\hline
\( \mathcal{B}_i^{t'} \) & Bounding boxes around objects relevant to the question in frame \( V_i^{t'} \), generated by MiniGPT-4. \\
\hline
\( F_o^{t'} \) & Region of Interest (RoI)-aligned object appearance features for each object in frame \( V_i^{t'} \). \\
\hline
\( F_s^{t'} \) & Spatial locations of objects in frame \( V_i^{t'} \). \\
\hline
\( F_I^{t'} \) & Frame-level feature representing the overall context of the frame \( V_i^{t'} \). \\
\hline
\( \mathcal{G}_i^{t'} \) & Frame-specific graph for frame \( V_i^{t'} \), constructed using object bounding boxes and frame context. \\
\hline
\( A^{t'} \) & Node set for the frame-specific graph \( \mathcal{G}_i^{t'} \). \\
\hline
\( R^{t'} \) & Edge weights in the frame-specific graph \( \mathcal{G}_i^{t'} \), calculated using self-attention on object features. \\
\hline
\( \hat{\mathbf{Q}} \) & Masked question embedding used in the Q-DGT module. \\
\hline
\( \mathcal{F}_{local}^{t'} \) & Local representation for frame \( V_i^{t'} \), derived from the Q-DGT module. \\
\hline
\( \mathcal{F}_{global} \) & Global video representation, aggregating spatial and temporal representations from all frames. \\
\hline
\( Z_{\hat{\mathbf{Q}}} \) & Textual embeddings of the question, projected into the textual information space. \\
\hline
\( \mathcal{F}_{final} \) & Final video representation, obtained by merging global and local representations using cross-attention. \\
\hline
\( \hat{A} \) & Predicted answer for the question, based on similarity between \( \mathcal{F}_{final} \) and pre-encoded answer representations. \\
\hline
\end{tabular}
\caption{Table of variables and descriptions used in the LGQAVE framework.}
\label{tab:variables}
\end{table*}
{\small
\bibliographystyle{ieee_fullname}
\bibliography{egbib}
}